\documentclass[a4paper, 11pt]{article}
\usepackage[T1]{fontenc}
\usepackage{amsmath, amsthm, amssymb, bm, pifont, dsfont}
\usepackage{stmaryrd}
\usepackage{nameref}
\usepackage[english]{babel}
\usepackage[authoryear]{natbib}
\bibpunct[; ]{(}{)}{,}{a}{,}{;}

\usepackage{graphicx}
\usepackage[table]{xcolor}
\usepackage{listings}
\usepackage{stmaryrd}
\usepackage{algorithm}
\usepackage{algpseudocode}
\usepackage{hhline}
\usepackage{multirow}
\usepackage{csvsimple}
\usepackage[font=small,skip=1pt]{caption}

\definecolor{Blue}{rgb}{0.1, 0.1, 0.5}
\usepackage{hyperref}
\hypersetup{ bookmarksnumbered, colorlinks=true, linkcolor=black,
  citecolor=Blue, breaklinks, pdfborder=0 0 0, pdfauthor={Aurélien Pion,
  Emmanuel Vazquez}, pdfsubject={Calibration of Gaussian process models},
  pdftitle={Calibration of Gaussian process models}, pdfkeywords={Gaussian
  processes, calibration, predictive distributions}}
\usepackage{color}

\renewcommand \hat    \widehat

\renewcommand \mathbb {\mathds}  

\newcommand   \Dcal   {\mathcal{D}}

\newcommand   \Ncal   {\mathcal{N}}

\newcommand   \E      {\mathsf{E}}
\newcommand   \R      {\mathbb{R}}

\newcommand   \N      {\mathbb{N}}

\newcommand   \X      {\mathbb{X}}
\renewcommand \P      {\mathsf{P}}
\newcommand   {\one}  {\mathds{1}}

\DeclareMathOperator \GP {{\rm GP}} \DeclareMathOperator \var {{\rm var}}

\newcommand{\ntrain}[1][n]{#1^{\text{train}}}

\newcommand{\xtest}[1][i]{x^{\text{test}}_{#1}}

\newcommand{\Ztest}[1][i]{Z^{\text{test}}_{#1}}
\newcommand{\ntest}[1][n]{#1^{\text{test}}}

\providecommand \abs [1]  {\lvert#1\rvert}

\newcommand\blfootnote[1]{%
  \begingroup
  \renewcommand\thefootnote{}\footnote{#1}%
  \addtocounter{footnote}{-1}%
  \endgroup
}

\begin{document}
\title{Gaussian process interpolation \\
  with conformal prediction: \\
  methods and comparative analysis}

\author{Aurélien Pion$\mbox{}^{1,\,2}$ and Emmanuel Vazquez$\mbox{}^{1}$ \\[1em]
\small
Universit\'e Paris-Saclay, CNRS, CentraleSup\'elec,\\
\small Laboratoire des Signaux et Syst\`emes, Gif-sur-Yvette, France \\
\small \texttt{firstname.lastname@centralesupelec.fr} \\
\small Transvalor S.A., Nice,  France}
\maketitle
{\bf Abstract:}  This article advocates the use of conformal prediction (CP) methods
  for Gaussian process (GP) interpolation to enhance the calibration of
  prediction intervals. We begin by illustrating that using a GP model
  with parameters selected by maximum likelihood often results in
  predictions that are not optimally calibrated. CP methods can adjust
  the prediction intervals, leading to better uncertainty quantification
  while maintaining the accuracy of the underlying GP model. We compare
  different CP variants and introduce a novel variant based on an
  asymmetric score. Our numerical experiments demonstrate the
  effectiveness of CP methods in improving calibration without
  compromising accuracy. This work aims to facilitate the adoption of CP
  methods in the GP community.

{\bf keywords} Gaussian processes; Prediction intervals;
    Calibration; Conformal Prediction

\section{Introduction}
\blfootnote{This work was supported by Transvalor S.A.}

Building an approxi\-mation---whether interpolation or
regression---of a computer code represented by a function
$f: \X \to \R$, where $\X \subseteq \R^d$ and $d \in \mathbb{N}^*$,
allows for predicting the result of the possibly expensive evaluation of
$f$ at a given $x\in\X$. It is often necessary to estimate the
uncertainty resulting from the approximation. Using the framework of
Gaussian Processes (GP) is a standard Bayesian approach to build such
approximations, together with predictive distributions that quantify
uncertainty.

In the following, we focus on interpolation and we consider a prior on
$f$ under the form of a GP $Z \sim \text{GP}(m, k)$, where
$m: \X \to \R$ is a mean function and $k: \X \times \X \to \R$ is a
positive definite covariance function. Given data
$\Dcal_n = \{(x_1, Z_1), \ldots, (x_n, Z_n)\}$, with $Z_i = Z(x_i)$,
$i = 1, \ldots, n$, the posterior mean function
$m_n(\cdot)=\E(Z(\cdot) \mid \Dcal_n)$ provides an approximation of the
underlying function and the posterior variance
$\sigma_n^2(\cdot)= \var(Z(\cdot)\mid\Dcal_n)$ offers a measure of
confidence in the approximation
\citep[see, e.g.,][]{rasmussen_gaussian_2006,stein_interpolation_1999}. 
This allows
for the construction of prediction intervals and the identification of
regions where uncertainty is high, which is often used for guiding further
data collection and model refinement, as in Bayesian optimization
\citep[see, e.g.,][]{feliot2017bayesian}.

However, in GP interpolation, the posterior variance does not depend on
the $Z_i$s but only on $m$, $k$ and the $x_i$s. To make
the posterior variance depend on the data, which is obviously desirable,
parameterized processes $Z$ are considered, with a parameter $\theta$
that is commonly selected by maximum likelihood (ML) or cross-validation
techniques, thus making it dependent on the observed data. Numerical
findings by \citet{petit_parameter_2023} show that ML generally yields
good predictive distributions when the covariance is an
anisotropic Matérn function. Additionally,
\citet{karvonen_maximum_2020} show that ML predictions are generally not
too optimistic, in the sense that the posterior variance does not
decrease too quickly as the number of observations increases over a
fixed domain.

Consider prediction intervals $I_{n, \alpha}(x)$ at level
$\alpha\in \left[ 0,\,1 \right[$, for the approximation of
$Z$ at $x\in\X$ from $\Dcal_n$.  Typically, such intervals can be derived
from the posterior distributions within the Bayesian framework:
\begin{equation}
    \label{eq:gp_int}
    I_{n,\,\alpha}(x) = \bigl[ m_n(x) + \Phi^{-1}( (1-\alpha)/2)\,{\sigma}_n(x),~ m_n(x) + \Phi^{-1}((1+\alpha)/2)\,{\sigma}_n(x)\bigr],
\end{equation}
where $\Phi^{-1}$ stands for the quantile function of the normal inverse
distribution. For any $x\in\X$, we have
$\P_n(Z(x)\in I_{n,\,\alpha}) = \alpha$, where $\P_n$ stands for the
conditional probability given $\Dcal_n$. For any $\alpha$, the empirical coverage
$$
\delta_{n,\,\alpha} = \frac{1}{\ntest} \sum_{i=1}^{\ntest} \one_{\Ztest[i] \in
  I_{n, \alpha}(\xtest[i])}
$$
computed on $\ntest$ points $(\xtest[i], \Ztest[i])$, with
$\Ztest[i]=Z(\xtest[i])$, should ideally be close to~$\alpha$, as this
is a desirable property expected by users. In this case, we say that
the prediction intervals are well calibrated.

\begin{figure}[h!]
  \centering
  \includegraphics[width=1.0\textwidth]{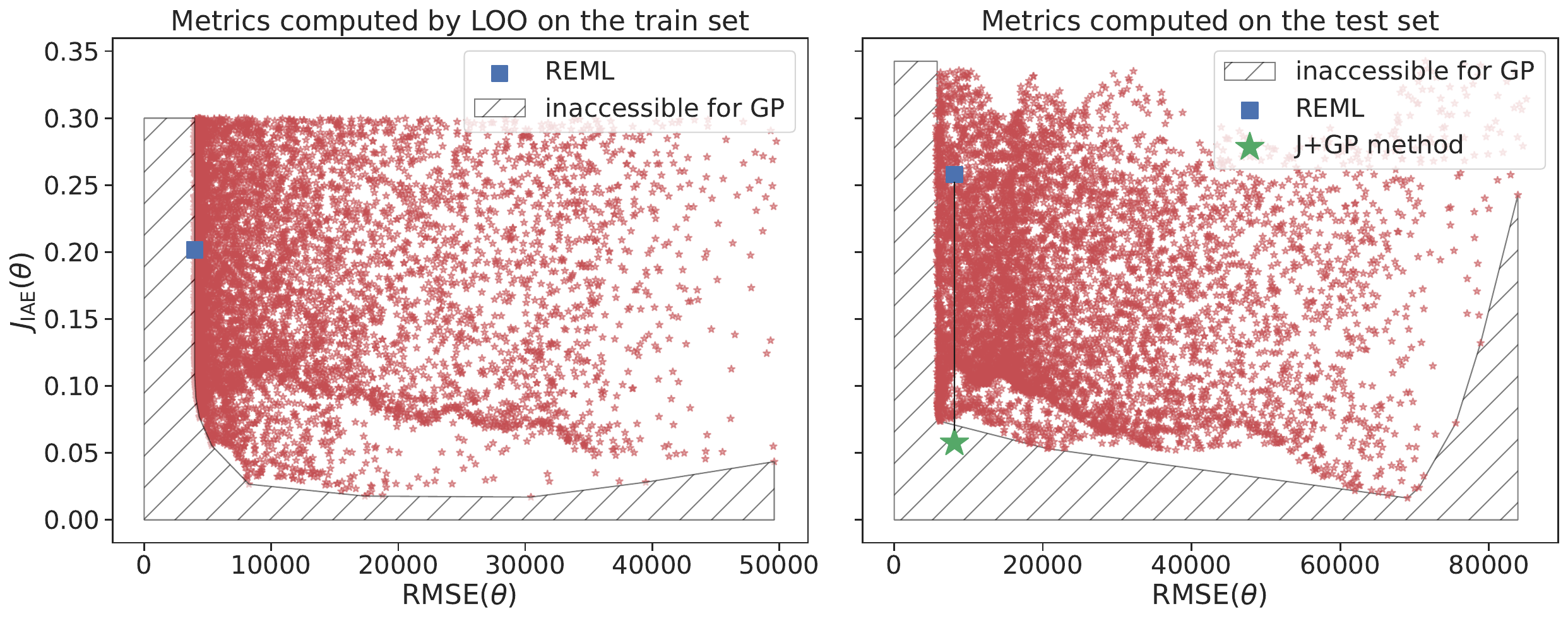}
  \caption{IAE versus RMSE, computed by leave-one-out (LOO) on the left-hand
    side, and on a test set of $1500$ points on the right-hand
    side. Each red point
    correspond to a different value for the variance and the range/lenghscale parameters of
    the covariance of $Z$. The blue square represents the parameter selected by
    REML, and the green star represents the metrics computed when a conformal
    prediction method (Jackknife+ for GP) is used to build the prediction
    intervals. Values, in LOO, above $J_{\rm IAE} > 0.3$ and RMSE $> 5\cdot10^4$ are not shown.}
  \label{fig:pareto}
\end{figure}

However, particularly when the model is misspecified, ML may yield overly
optimistic or overly pessimistic predictions (too small or too large empirical
coverage).  This is illustrated in Figure~\ref{fig:pareto}, which shows a
scatter plot of the values of the integrated absolute error~\citep[IAE,][]{marrel_algo}
$$
J_{\rm IAE}(\theta) = \int_{0}^1 |\delta_{\alpha} - \alpha|{\rm d}\alpha \,,
$$
that quantifies the calibration of the predictive distributions, versus the
root-mean-square error (RMSE) for a given $\theta$. Here, the data
$\mathcal{D}_n$ is obtained using one draw of a uniform distribution in $\X$,
with $n=150$ points, evaluated on the Goldstein-Price test function
\citep[$d=2$ ; see, e.g.,][]{simulationlib}. The GP model $Z$ has an unknown constant mean
and an anisotropic Matérn covariance function \citep{stein_interpolation_1999},
with regularity parameter $\nu = 2.5$. Each point corresponds to a random value
for the parameter $\theta$ of the covariance (variance at origin, range/lengthscale
parameters). On the left-hand side, the metrics (RMSE, IAE) are computed using
$\mathcal{D}_n$ only and a leave-one-out (LOO) strategy is used, while on the right-hand
side, they are computed on a test set of $1500$ points.

The blue square corresponds to the parameter selected by restricted-ML (REML)
\citep{stein_interpolation_1999}.  Notice that on the left hand-side, the REML
point minimizes the RMSE as noted by~\citet{petit_parameter_2023}.
On the test set, the REML point does not minize the RMSE but stays nevertheless close to
the
Pareto front. No red points reach a value of zero for the IAE,
meaning that no parameter of the GP model yields perfectly calibrated predictive
distributions. While being a particular case, and a particular function,
this situation is by no means exceptional. The code to
reproduce this experiment is available to the
reader\footnote{See \url{https://github.com/gpmp-dev/lod2024-conformal}}.

As shown in Figure~\ref{fig:pareto}, it turns out that it is possible to improve
the calibration of the prediction intervals using \textit{conformal prediction}
(CP). CP is a framework originally proposed by \citet{vovk_Gammerman_2005} to construct prediction intervals for any
prediction algorithm, with known statistical properties, under certain
assumptions. Recent developments have extended the original method to make it
more suitable for Gaussian process approximation
\citep{jaber_conformal_2024, lei_distribution_free_2018}. Thus, in
Figure~\ref{fig:pareto}, the green star point represents the metrics computed
when the Jackknife+ for GP (J+GP) method of~\citet{jaber_conformal_2024} is used to construct the prediction intervals.
IAE is now closer to zero with the same RMSE, and the calibration has been
improved significantly without reducing the accuracy.

The first objective of this article is to advocate for CP-type methods for GP
approximation to encourage their adoption by the GP community. Compared to
existing works, particularly the work of~\citet{jaber_conformal_2024}, this article
specifically examines the noise-free case, that is, GP interpolation,
whereas~\citet{jaber_conformal_2024} only consider the case with
observation noise, i.e., GP regression. Additionally, our numerical
experiments cover a larger number of functions. We also include in our
comparisons the full conformal prediction for GPs proposed
by~\citet{papadopoulos_guaranteed_2023}, as well as a new variant based
on an asymmetric score.

The article is organized as follows. In Section 2, we recall the main
ideas of CP. Section 3 shows how to apply CP to Gaussian process
approximation. In Section 4, we conduct numerical experiments to compare
different variants of CP in the case of GP interpolation, using several test functions.

\section{Conformal prediction for regression}

In this section, we briefly recall the principles of conformal
prediction for regression. The reader is referred to
\cite{lei_distribution_free_2018,  barber_predictive_2021} for
more details.

\subsection{Non-conformity scores}

Consider i.i.d. data $(X_i, Z_i)$,
$i=1,\,2,\ldots$, from a common distribution on $\X \times \R$, where
the $X_i$s are covariates and the $Z_i$ are the corresponding responses,
with mean $\E(Z_i \mid X_i) = f(X_i)$. We assume that a method is
available for constructing  regression functions, denoted by
$s(\,\cdot\,;\,\mathcal{D})$, that estimates $f$ from
a finite dataset $\mathcal{D} = \left\{ (X_i, Z_i), i=1,\,2,\ldots \right\}$.

Conformal prediction aims at producing a prediction interval
$I_{n, \alpha}(X)$ for a new response $Z$ at $X$, given $\mathcal{D}$.
The main idea of CP is to evaluate how well a potential value $z$ for
$Z$ fits with the existing dataset when added as a new data
point. This is done by introducing a non-conformity score, $R$, which
measures the "distance" or non-conformity of an observation
$(x,\, z) \in \X\times\R$ with respect to the dataset $\Dcal$.
A common choice for the non-conformity score is the residual error
$$
R(x,\, z\,;\, \mathcal{D}) = \abs{z  -  s(x \, ; \,\mathcal{D})}.
$$

\subsection{Full conformal prediction}

Let $\mathcal{D}_n = \left\{ (X_i, Z_i), i=1,\,\ldots,\,n \right\}$.
For a new random covariate $X_{n+1}$ and a potential value $z\in\R$ for the
response $Z_{n+1}$, define the augmented
dataset
$$
\mathcal{D}_{n+1, z} = \{(X_1, Z_1), \ldots, (X_n, Z_n),
(X_{n+1}, z)\}.
$$
Consider the scores
\begin{equation}
  \label{eq:fcp-scores}
   R_{z, i} = R(X_{i}, Z_i\,;\, \mathcal{D}_{n+1, z}),~ i=1,\ldots,\,n, \text{ and } R_{z, n+1} = R(X_{n+1}, z\,;\, \mathcal{D}_{n+1,
  z}).
\end{equation}
The original CP method of \citet{vovk_Gammerman_2005}, also called full
conformal prediction (FCP),
consists in building the prediction interval at $X_{n+1}$
defined by 
\begin{equation}
    \label{eq:fcp-ci}
    I_{n,\, \alpha}^{\rm FCP} (X_{n+1}) = \bigl\{ z  \in \R, \gamma(z) \leq \left\lceil \alpha(n+1) \right\rceil \bigr\}\,,
\end{equation}
where
\begin{equation}
  \label{eq:fcp-gamma}
  \gamma(z) = \sum_{i=1}^{n+1} \one_{R_{i, z} \leq R_{n+1, z}}
\end{equation}
is the number of non-conformity scores less than or equal to
$R_{n+1, z}$.

The interval $I_{n,\,\alpha}^{\rm FCP}$ verifies the
finite-sample coverage property:
$\P(Z_{n+1}\in I_{n, \alpha}^{\rm FCP}(X_{n+1})) \geq \alpha$ 
\citep[see][Theorem 2.1]{lei_distribution_free_2018}, where the probability
is taken over the $n+1$ i.i.d. draws $(X_1, Z_1),\,\ldots,\,(X_{n+1}, Z_{n+1})$. While FCP provides
statistically calibrated intervals, it can be computationally expensive
since it requires testing several potential values $z$ to get (an
approximation of) $I_{n,\, \alpha}^{\rm FCP} (X_{n+1})$. For each
candidate $z$, one needs to recalculate the non-conformity scores, which
requires fitting a new model. This process results in significant
computational overhead, especially for large datasets or complex
models. This limitation has led to the development of alternative
approaches, such as split conformal prediction, as recalled next.

\subsection{Split conformal prediction}

An alternative approach is split conformal prediction (SCP), which partitions
the data into two sets: a training set and a calibration set. The
training set is used to fit the regression model, while the calibration
set is used to compute the non-conformity scores: given a partition of the data
$\Dcal_n = \Dcal^{\rm train} \cup \Dcal^{\rm cal}$, fit the
regression model on $\mathcal{D}^{\rm train}$ and use
$\mathcal{D}^{\rm cal}$ to compute the non-conformity scores.

The prediction interval for a new response $Z_{n+1}$ at $X_{n+1}$ is then given
by
\begin{equation}
    \label{eq:scp-ci}
    I_{n, \alpha}^{\rm SCP} (X_{n+1}) = \left\{ z \in \R,~ R(X_{n+1}, z\,;\, \mathcal{D}^{\rm train}) \leq q_{\alpha} \right\}\,,
\end{equation}
where $q_{\alpha}$ is the $\alpha$ quantile of the set $\{ R(X_i, Z_i\,;\,
\mathcal{D}^{\rm train}),~(X_i, Z_i)\in \mathcal{D}^{\rm cal}\}$ of
non-conformity scores from the calibration set.

This approach provides a computationally efficient way to construct prediction
intervals while maintaining statistically valid coverage. However, the choice of
the split between the training and calibration sets can significantly affect the
tightness and coverage of the prediction intervals because each split uses fewer
data for training and calibration. To address these limitations, alternative
methods such as Jackknife conformal prediction have been developed. These
methods aim to use the entire dataset more effectively, reducing the dependency
on a single partition.

\subsection{Jackknife conformal prediction}

Jacknife conformal prediction (JCP) uses all data points both for training and
calibration by employing a LOO strategy, where,
for each data point $(X_i, Z_i)$ in $\mathcal{D}_n$, the
regression model is fitted on the dataset 
$\Dcal_{n,\,-i} = \Dcal_n \setminus \{(X_i, Z_i)\}$.

The prediction interval for a given point $x\in\X$ is then defined by
\begin{equation}
    \label{eq:jcp_set}
    I_{n, \alpha}^{\rm JCP} (x) = \left[ s(x; \Dcal_{n}) - q_{\alpha},~s(x; \Dcal_{n}) + q_{\alpha}\right] 
\end{equation}
where $q_{\alpha}$ is the empirical $\alpha$-quantile of the set of scores
$R(X_i, Z_i\,;\, \mathcal{D}_{n,\,-i})$, for $i=1,\,\ldots,\,n$.

Contrarily to FCP and SCP, JCP does not provide strong theoretical guarantees,
but has nevertheless the (weak) following finite-sample in-sample property:
$\P(Z_i\in I_{n,\,\alpha}^{\rm JCP}(X_i)) \geq \alpha$, for all
$i=1,\,\ldots,\,n$. If the dataset is large enough, we can expect this property
to extend to a new, unobserved point \citep[see][Section~4]{barber_predictive_2021}. Moreover, since JCP uses the entire
dataset for both model fitting and calibration, JCP can often produce tighter
intervals than those obtained with SCP.

\subsection{Jackknife+ conformal prediction}

While JCP method has the advantage of using the entire dataset for both
training and calibration, it does not provide strong theoretical
guarantees as mentioned above. To address this, \citet{barber_predictive_2021} introduced the Jackknife+ 
method (J+), which enhances JCP to provide prediction intervals with
finite-sample coverage properties.

Barber et al. observe that regression algorithms can be sensitive to
the specific dataset used for training, causing the predictions from
models trained on slightly different datasets (e.g.,
$\Dcal_{n, -i}$) to vary significantly. To mitigate this issue,
J+ modifies JCP by considering two sequences,
\begin{equation}
  \label{eq:j+-seqs}
\xi_i^+ = s(X_{n+1}; \Dcal_{n, -i}) + R_i \text{ and } \xi_i^- = s(X_{n+1}; \Dcal_{n, -i}) - R_i,
\end{equation}
where $R_i = R(X_i, Z_i\,;\, \Dcal_{n, -i})$, $i=1,\,\ldots,\,n$.

The sequences $\{\xi_i^-\}_{i=1}^n$ and $\{\xi_i^+\}_{i=1}^n$ are then
ordered, and the J+ confidence interval at $X_{n+1}$ is defined by
\begin{equation}
  \label{eq:j+-ci}
  I_{\alpha}^{\rm J+}(X_{n+1}) = \left[\xi_{(\lfloor(n+1) (1-\alpha)\rfloor)}^-,~\xi^+_{(\lceil(n+1)\alpha\rceil)}\right],
\end{equation}
provided $\alpha \leq n/(n+1)$.

The J+ method verifies the finite-sample coverage property
$$
\P\bigl(Z_{n+1} \in I_{n, \alpha}^{\rm J+}(X_{n+1})\bigr) \geq 2\alpha - 1
$$
under the i.i.d. assumption for  $(X_1, Z_1), \ldots, (X_{n+1}, Z_{n+1})$
\citep[see][Theorem~1]{barber_predictive_2021}, but typically achieves
coverages close to $\alpha$ in practice.

\section{Conformal prediction for Gaussian processes}

In this section, we return to the case of GP interpolation, as presented
earlier in this article. More precisely, we consider an unknown function
$f$ and aim to build an approximation using the framework of GP
interpolation. To this end, we assume a GP prior $Z \sim \GP(m, k)$ with
known mean and covariance functions, typically determined through a model
selection procedure such as maximum likelihood or cross-validation techniques.
Given a dataset $\Dcal_n = \{(x_1, Z_1), \ldots, (x_n, Z_n)\}$, where $Z_i =
Z(x_i)$ for $i = 1, \ldots, n$, GP interpolation consists in computing the
posterior distribution of $Z(x)$ for all $x\in \X$. We now briefly present the
adaptation of CP to GP interpolation. Subsequently, we introduce a novel
non-conformity score for the Jackknife+ method.

\subsection{Adaptation of Full-Conformal Prediction and Jackknife+}

\subsubsection{Full-Conformal Prediction for GP (FCP-GP) ---}
FCP is adapted by \citet{papadopoulos_guaranteed_2023} to GP
interpolation, building on the earlier adaptation of CP to kernel ridge
regression by \citet{ridge_regression}. The main idea involves
rewriting the non-conformity scores~\eqref{eq:fcp-scores}.

Specifically, given a
dataset $\Dcal_{n}$, and a new point $x\in\X$, define the augmented
dataset
$$
\Dcal_{n+1} = \Dcal_{n} \cup \{(x, Z_{n+1})\}\,,\quad Z_{n+1}= Z(x),
$$
assuming that all points $x_i$ and $x$ are distinct.
Denote by $m_{n+1, -i}(x)$ and $\sigma^2_{n+1, -i}(x)$ the posterior
mean and variance of $Z(x)$ from the LOO dataset
$\Dcal_{n+1,\,-i} = \Dcal_{n+1}\setminus \left\{ (x_i, Z_i)
\right\}$,
$i=1,\,\ldots,\,n+1$. Then, consider the scores
\begin{equation}
  \label{eq:scores-gp}
  R_i = \frac{\abs{Z_i - m_{n+1, -i}(x_i)}}{\max (\epsilon,
    \sigma_{n+1, -i}^{\beta}(x_i))}, \quad i=1,\,\ldots,\,n+1,
\end{equation}
where $\beta > 0$ is a parameter that controls the sensitivity to
changes in the variance $\sigma_{n+1, -i}^2(x_i)$, and $\epsilon \geq 0$
is a small constant introduced to ensure numerical stability when
$\sigma_{n+1, -i}^2(x_i)$ is small. Taking $\beta = 1$ is a sensible
choice, since under our Bayesian setting, the random variables
$(Z_i - m_{n+1, -i}(x_i))/\sigma_{n+1, -i}(x_i)$ are $\Ncal(0,\,1)$ and
do not depend on the parameters of the GP model.

The FCP-GP method replaces the definition of $\gamma$ in (\ref{eq:fcp-gamma}) by
\begin{equation}
  \label{eq:fcp-gp-gamma}
  \gamma(z) = \E \Bigl[ \sum_{i=1}^{n+1} \one_{R_i \leq R_{n+1}} \mid Z_{n+1} = z \Bigr]
\end{equation}
and computes the prediction intervals as in~(\ref{eq:fcp-ci}). Using the
linearity of $m_{n+1, -i}$ with respect to $Z_{n+1}$,
\citet{papadopoulos_guaranteed_2023} shows that these intervals can be computed
efficiently. Note also that \citet{papadopoulos_guaranteed_2023}
proposes other types of scores that we do not present here.

\subsubsection{Jackknife+ for GP (J+GP) ---}
Using the scores $R_{i}$ defined by~(\ref{eq:scores-gp}) computed with
the training dataset $\mathcal{D}_n$ instead of the augmented dataset
$\Dcal_{n+1}$, \citet{jaber_conformal_2024} propose another
CP method for GP approximation based on a modification of the J+ method.
The sequences~(\ref{eq:j+-seqs}) are replaced by
\begin{align}
  \label{eq:j+gp-seqs}
  \xi_i^+(x) &= m_{n, -i}(x) + R_i \max \left(\epsilon,
    \sigma_{n, -i}^{\beta}(x)\right)\,, \\
  \xi_i^-(x) &= m_{n, -i}(x) - R_i \max \left(\epsilon, \sigma_{n, -i}^{\beta}(x)\right),
\end{align}
and the prediction intervals are computed
using~(\ref{eq:j+-ci}). Intervals calculated in this way have the same
coverage property as that of the J+ method for i.i.d. data
$(X_1, Z_1), \ldots, (X_{n+1}, Z_{n+1})$, conditional on $Z$---as when
the $X_i$ are uniformly distributed on $\X$.

\medskip

\noindent
{\bf Remark.} Score normalization has been suggested by \citet{vovk_Gammerman_2005}, Chapter 2.3, in their application of
full-conformal to ridge regression to adapt to noisy observations. They choose a
normalization coefficient such that all the non-conformity scores have
relatively the same variance. Later, \citet{lei_distribution_free_2018} discussed the scaling of residual errors
$\abs{z - s(z, \Dcal)}$ in the case of heteroscedastic noise on the
observations, by a measure of the local dispersion.

\subsection{Asymmetric scores}

\citet{barber_predictive_2021} advocate using signed scores
when dealing with skewed data. If the data exhibit asymmetry, such as local
excursions, the prediction intervals should ideally reflect this asymmetry. To
this end, we introduce a modification of the scores~(\ref{eq:scores-gp}) by
removing the absolute value:
\begin{equation}
  R_i = \frac{Z_i - m_{n, -i}(x_i)}{\max (\epsilon, {\sigma}_{n, -i}(x_i))}.
\end{equation}
Then, define the sequence
$$
\xi_i(x) = m_{n, -i}(x) + R_i \max \left(\epsilon,  \sigma_{n, -i}(x)\right).
$$
The prediction intervals for a given $x\in \X$ is obtained as
\begin{equation}
    I_{n,\alpha}^{\rm asymJ+GP}(x) = \left[\xi_{\left\lfloor (1 - \alpha) / 2 (n+1)\right\rfloor}(x)\,,~ \xi_{\left\lfloor (1 + \alpha) / 2 (n+1)\right\rfloor}(x)\right].
\end{equation}
  
We refer to this method as asymJ+GP. Moreover, the finite-sample coverage
property for i.i.d. data, $\P\bigl(Z_{n+1} \in I_{\alpha}^{\rm
J+}(X_{n+1})\bigr) \geq 2\alpha - 1$, remains valid, as demonstrated in Appendix
A of \cite{barber_predictive_2021}.

\section{Numerical Comparison}
\label{sec:num}

In this section, we conduct a numerical comparison between the different CP
methods. We use GPmp, the Python GP micro
package~\citep{gpmp}, for the implementation.

We consider  GP models $Z$ with a constant mean and a Matérn
covariance function with regularity $\nu = p + 1/2$, where $p \in \N^\star$. The
Matérn correlation structure $\kappa_{\nu}$ is defined in Chapter 2.7 of
\cite{stein_interpolation_1999}, and the corresponding anisotropic
covariance function can be written as
\begin{equation}
\label{eq:matern-cov}
k_{\sigma, \nu, \rho} (x, y ) = \sigma^2 \kappa_{\nu}\left( \sqrt{ \sum_{i=1}^d
  \frac{(x_{[i]} -y_{[i]})^2}{\rho_i^2}}\right) \qquad x,\, y\in\R^d\,.
\end{equation}
The parameter $\nu$ controls the regularity of the covariance, $\sigma^2$ is a
variance parameter, and the parameters $\rho_i$ are lengthscale parameters
controlling the scale of variations along each dimension. The parameters
$\sigma^2$ and the $\rho_{i}$s are selected by REML.

The goal is to understand the behavior of the prediction intervals computed
either using~(\ref{eq:gp_int}) or a CP method when the GP model has different
and not necessarily ``optimal'' values of regularity $\nu$.

For a given value of $\nu$, we sample $\ntrain = 20 \times d$ points in $\X$,
$(x_1, \ldots, x_{\ntrain})$, using the uniform distribution on $\X$, and
compute the outputs $f(x_1)$, \ldots, $f(x_{\ntrain})$. We apply the same
strategy to compute the test set with $\ntest = 1100$ points. This procedure is
repeated $40$ times to compute on each repetition the empirical coverage
$\delta_{\alpha}$, with $\alpha = 90\%$, the spatial average of the width of the
intervals, and the IAE on the test set. All non-conformity scores are computed
with $\beta = 1$.

The test functions used for experiments are the Goldstein-Price function ($d =
2$), the Hartmann4 function ($d = 4)$, the Hartmann6 function ($d = 6$), the
Park function ($d = 4$), the Branin function ($d=2$), and a Becker function in
dimension $d = 2$. Information about the Goldstein-Price, and Hartmann functions
can be found in~\cite{simulationlib} and information about the Branin function
can be found in~\cite{dixon}. The Park function is defined in, e.g.,
\cite{park}. The Becker functions are from~\cite{becker_metafunctions_2020}.

\begin{figure}[h!]
   \centering
   \includegraphics[width=\textwidth]{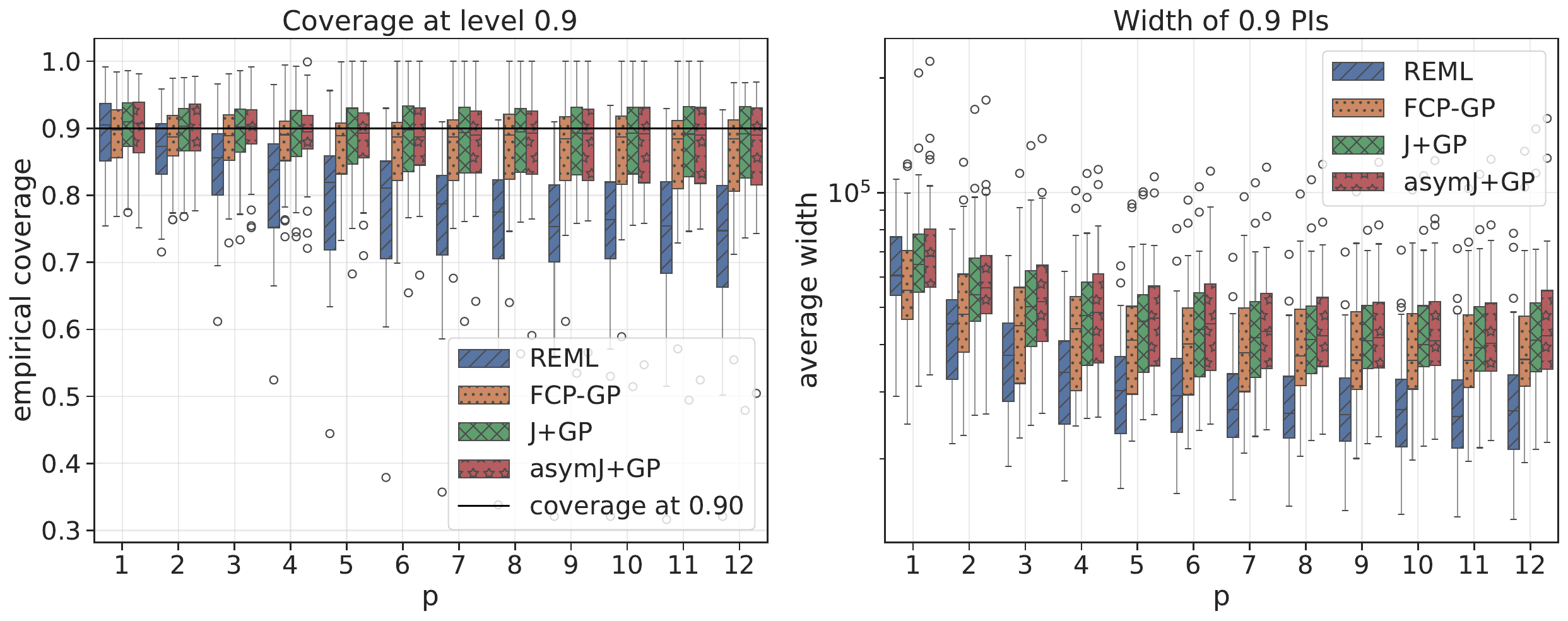}
   \caption{Coverage and average width of the prediction intervals at level $0.9$
     for the Goldstein-Price function with $40$ training points. The GP
     model parameters $\sigma$ and $\rho$ are selected by REML. The
     intervals are computed using the posterior variance, the 
     FCP-GP method, the J+GP method, and the asymJ+GP method.}
   \label{fig:goldstein_price}
\end{figure}

Figure~\ref{fig:goldstein_price} displays the boxplots of the empirical
coverage and the average width of the intervals (over the $40$
repetitions) when the regularity parameter varies, in the case of the
Goldstein-Price function.  All CP methods give better coverage than the
GP model when the parameters are selected by REML, which is generally
overconfident when $p$ increases. The FCP-GP method is more optimistic
(smaller prediction intervals, smaller coverage) than J+GP or
asymJ+GP. The improvement in coverage by the CP methods is achieved by
increasing the size of the prediction interval, as shown by the
right-hand side of Figure~\ref{fig:goldstein_price}.

Table~\ref{tab:res} summarizes the performance of the methods for the
other test functions. The average IAE and the average width of the
prediction interval at the 90\% level, computed on the test set, are
reported for several values of model regularity $\nu = p + 1/2$. The IAE
reflects the coverage at multiple levels, and all CP methods improve the
IAE compared to REML. All CP methods produce relatively similar
results. The asymJ+GP method gives very similar results to J+GP and has
a better IAE for the Beck function. However, it should be noted that the
asymJ+GP method produces wider prediction intervals on average than
J+GP.

\begin{table}[!h]
  \centering
  \setlength\tabcolsep{4pt}
  \begin{tabular}{l lcccccccc}
  \hline
  \multirow{2}{*}{\textbf{Function}} & \multirow{2}{*}{\textbf{p}} & \multicolumn{2}{c}{\textbf{REML}} & \multicolumn{2}{c}{\textbf{FCP-GP}} &
  \multicolumn{2}{c}{\textbf{J+GP}} & \multicolumn{2}{c}{\textbf{asymJ+GP}} \\
  ~ & ~ & 90\%W & IAE & 90\%W & IAE & 90\%W & IAE & 90\%W &
  IAE \\ \hline \hline
   & \textbf{1} & 0.045 & 0.21 & 0.035 & 0.06 & 0.043 & 0.06 & 0.047 & 0.06 \\
  \scriptsize Beck  & 5 & 0.0059 & 0.1 & 0.009 & 0.09 & 0.01 & 0.08 & 0.011 & \textbf{0.07} \\
   & 9 & 0.0064 & 0.24 & 0.012 & 0.12 & 0.013 & 0.11 & 0.014 & 0.11 \\
  \cline{1-10}
  \rowcolor{lightgray} & \textbf{1} & $1.4\cdot 10^1$ & 0.24 & 7.6 & 0.06 & 8.7 & 0.06 & 9.1 & 0.06 \\
  \rowcolor{lightgray} \scriptsize Branin & \textbf{5} & 0.52 & 0.12 & 0.5 & 0.1 & 0.51 & 0.1 & 0.54 & 0.1 \\
  \rowcolor{lightgray} & \textbf{9} & 0.75 & 0.08 & 0.9 & 0.09 & 0.88 & 0.09 & 0.91 & 0.09 \\
  \cline{1-10}
   & \textbf{1} & $6.4\cdot 10^4$ & 0.19 & $6\cdot 10^4$ & 0.06 & $7\cdot 10^4$ & 0.06 & $7.4\cdot 10^4$ & 0.06 \\
   \scriptsize Goldstein Price& \textbf{5} & $3.3\cdot 10^4$ & 0.09 & $4.4\cdot 10^4$ & 0.08 & $4.8\cdot 10^4$ & 0.08 & $4.9\cdot 10^4$ & 0.08 \\
  & \textbf{9} & $2.9\cdot 10^4$ & 0.11 & $4.2\cdot 10^4$ & 0.09 & $4.5\cdot 10^4$ & 0.08 & $4.6\cdot 10^4$ & 0.08 \\
  \cline{1-10}
  \rowcolor{lightgray} & \textbf{1} & 0.98 & 0.08 & 0.86 & 0.04 & 0.9 & 0.04 & 0.92 & 0.04 \\
  \rowcolor{lightgray} \scriptsize  Hartmann 4 & \textbf{5} & 0.84 & 0.05 & 0.82 & 0.04 & 0.83 & 0.04 & 0.85 & 0.04 \\
  \rowcolor{lightgray} & \textbf{9} & 0.83 & 0.05 & 0.82 & 0.05 & 0.83 & 0.04 & 0.85 & 0.04 \\
  \cline{1-10}
  & \textbf{1} & 0.68 & 0.14 & 0.54 & 0.03 & 0.56 & 0.03 & 0.6 & 0.03 \\
  \scriptsize Hartmann 6 & \textbf{5} & 0.64 & 0.11 & 0.54 & 0.04 & 0.56 & 0.04 & 0.59 & 0.04 \\
  & \textbf{9} & 0.63 & 0.11 & 0.53 & 0.04 & 0.56 & 0.04 & 0.59 & 0.04 \\
  \cline{1-10}
  \rowcolor{lightgray} & \textbf{1} & 0.072 & 0.25 & 0.032 & 0.04 & 0.034 & 0.04 & 0.034 & 0.05 \\
  \rowcolor{lightgray} \scriptsize Park Function & \textbf{5} & 0.0015 & 0.08 & 0.0015 & 0.07 & 0.0015 & 0.07 & 0.0015 & 0.07 \\
  \rowcolor{lightgray} & \textbf{9} & 0.0012 & 0.08 & 0.0013 & 0.08 & 0.0014 & 0.08 & 0.0014 & 0.08 \\
  \cline{1-10}
\end{tabular}
\caption{Average width of 90\% interval and average IAE for multiple test functions. 90\%W stands for the average width of 90\% prediction interval.}
\label{tab:res}
\end{table}

\section{Discussion}

CP methods can enhance prediction intervals, as shown in
Figure~\ref{fig:pareto}. All CP methods provide relatively similar
coverage, though J+GP tends to perform slightly better on average. For data with
excursions, such as the Goldstein-Price function, using
asymJ+GP should be preferable. In other scenarios, J+GP is generally a
good choice.

As a final remark, Figure~\ref{fig:pareto} suggests that the RMSE of the
GP model obtained by REML is near optimal, as already noted by
\citet{petit_parameter_2023}. Our findings indicate that it might be beneficial
to decouple the objectives of achieving high prediction accuracy (low
RMSE) and
obtaining reliable prediction intervals (low IAE), but we did not
explore the fully Bayesian approach
\citep[see, e.g.,][]{benassi11:_robus_gauss_proces_based_global} in this study.

\bibliographystyle{plainnat}
\bibliography{bibl} 

\end{document}